\documentclass[letterpaper,10pt,conference]{ieeeconf}
\IEEEoverridecommandlockouts    

\usepackage{multirow}
\usepackage{lipsum}
\usepackage{amsmath}
\usepackage{amssymb}
\usepackage[ruled,vlined]{algorithm2e}
\usepackage{graphicx}
\usepackage{subfig}
\usepackage{tabularx}
\usepackage{booktabs}
\usepackage[table]{xcolor}
\usepackage{array}
\usepackage{multirow}
\usepackage{xcolor}
\usepackage{graphicx,url}
\graphicspath{{./Figures/}}
\usepackage[T1]{fontenc}
\usepackage[utf8]{inputenc}
\usepackage{float}
\usepackage[hidelinks]{hyperref}

\definecolor{accent}{HTML}{2E75B6}

\title{\LARGE \bf Confidence-Controlled XAI Auditing for Pedestrian Detection under Domain Shift}

 \author{
	\parbox{\textwidth}{%
 		\centering
 		Ruben Dario Florez-Zela $^{1}$
 	}%
 	\thanks{$^{1}$Universidad Nacional de San Agustín de Arequipa, Arequipa, Peru {\tt\small (rflorezz@unsa.edu.pe)}}%
    \newline
    \thanks{\scriptsize
    © 2026 IEEE. Personal use of this material is permitted. Permission from IEEE must be obtained for all other uses, in any current or future media, including reprinting/republishing this material for advertising or promotional purposes, creating new collective works, for resale or redistribution to servers or lists, or reuse of any copyrighted component of this work in other works.}%
 }

\begin{document}
	
	\maketitle
	\thispagestyle{empty}
	\pagestyle{empty}	
	\begin{abstract}
    Explainability is increasingly required for \mbox{perception} models in intelligent vehicles, yet whether explanations remain faithful under driving domain shift is still poorly understood. This work audits post-hoc explanations of a fixed YOLOv8s pedestrian detector across PIE and JAAD using ROI-based D-Deletion, frozen confidence terciles, \mbox{rank-based} tests, bootstrap intervals, and Holm correction. The audit shows that deletion-based faithfulness is strongly coupled to detection strength at explanation time, with Spearman correlations between $0.70$ and $0.82$ for D-RISE, making naive confidence-stratified comparisons unreliable. After controlling for detection strength within fixed $f_0$ bins, D-RISE faithfulness remains domain-dependent in the central $f_0$ range, with PIE showing higher D-Deletion than JAAD and Holm-adjusted significance. A non-perturbative EigenCAM baseline is less faithful than D-RISE but also exhibits score coupling, suggesting that the effect is not specific to D-RISE and is related to the deletion-based evaluation setup. These results motivate confidence-controlled XAI audits for safety-critical perception under domain shift.     
    
    {\small\indent\textbf{\textit{Index Terms---}Explainable AI, pedestrian detection, domain shift, D-RISE, D-Deletion, intelligent vehicles.}}
    \end{abstract}

	\section{Introduction}
	\label{sec:introduction}
	
    Perceiving pedestrians and other vulnerable road users reliably is among the most safety-critical functions of an intelligent vehicle, since an error in this task translates directly into risk for the people least protected in traffic. Modern perception stacks address it with deep object detectors~\cite{redmon2016yolo}, and datasets recorded from a moving vehicle, such as PIE~\cite{rasouli2019pie} and JAAD~\cite{rasouli2017jaad}, have become common ground for developing and comparing pedestrian models. These detectors reach strong average accuracy, yet they remain opaque. A bounding box carries no account of which image evidence produced it, which is precisely the kind of information a safety case requires.

    Explainability has consequently become a requirement rather than a convenience for perception in automated driving, and a growing body of work frames it as a pillar of safe and trustworthy autonomy~\cite{atakishiyev2024xai,omeiza2022explanations}. Among post-hoc techniques, perturbation-based saliency is appealing because it treats the model as a black box and needs only its inputs and outputs. RISE established this idea for classifiers~\cite{petsiuk2018rise}, and D-RISE extended it to detectors through a similarity score that accounts for both localization and category~\cite{petsiuk2021drise}. An explanation is useful only if it is faithful, that is, if the regions it highlights are the ones the detector actually relies on. Faithfulness is commonly assessed with deletion curves, and recent work has adapted this idea to object detection, formalizing a deletion measure for detectors together with a dedicated explanation method~\cite{andres2024ddeletion}.

    What remains largely unexamined is whether these explanations, and the way their faithfulness is measured, survive a change of operating conditions. Faithfulness metrics for detectors have been validated mostly within a single dataset and outside the driving domain, so two questions stay open. First, does explanation faithfulness transfer across driving domains, or does it depend on where the detector is evaluated? Second, can faithfulness be compared across groups of detections without being confounded by the detector's own confidence, given that the safety-relevant cases are the \mbox{low-confidence} ones? Pedestrian detection across PIE and JAAD offers a natural setting to study both questions with a fixed detector and a fixed evaluation protocol, isolating the effect of the evaluation domain.

    The contributions are the following:
    \begin{itemize}
      \item A cross-domain auditing protocol for the faithfulness of pedestrian-detection explanations. The protocol evaluates a targeted perturbation-based explainer, D-RISE, against a non-perturbative CAM baseline, EigenCAM, under a common deletion metric on region-of-interest crops, fixes confidence terciles before any explanation is computed, and evaluates over multiple seeds with rank-based tests, bootstrap confidence intervals, and Holm correction.
      \item Evidence that, in this task, deletion-based faithfulness is coupled to the strength of the detection at explanation time, with Spearman correlations between $0.70$ and $0.82$. Reading the metric across confidence-defined groups is therefore confounded, a caution that extends to confidence-based evaluation of detector explanations more generally.
      \item Evidence that, once detection strength is controlled, explanation faithfulness still differs systematically between domains across the central $f_0$ range, which indicates that faithfulness is not invariant under domain shift.
    \end{itemize}

    \section{Related Work}

    Pedestrian detection is a safety-critical component of driver assistance and automated driving, and dedicated benchmarks have driven much of its progress~\cite{dollar2012ped,zhang2017citypersons}. Modern single-stage detectors such as YOLO~\cite{redmon2016yolo} provide the speed and accuracy required for on-board perception. Driving datasets such as PIE~\cite{rasouli2019pie} and JAAD~\cite{rasouli2017jaad} offer frame-level pedestrian annotations and are commonly used to study pedestrian behavior and detection. Prior work has shown that pedestrian detectors can lose accuracy across datasets because of differences in city, viewpoint, density, and occlusion~\cite{zhang2017citypersons}. However, this literature focuses on detection accuracy, not on whether the explanations of those detections remain faithful across domains.

    Post-hoc visual explanations include activation-based methods such as Grad-CAM~\cite{selvaraju2017gradcam} and perturbation-based saliency methods that treat the model as a black box. RISE introduced randomized masking for classifiers~\cite{petsiuk2018rise}, while D-RISE extended this idea to object detectors by scoring masked predictions with a target similarity that combines localization and category~\cite{petsiuk2021drise}. Faithfulness is commonly assessed with deletion curves, which remove salient pixels and measure the drop in model response; this idea has recently been adapted to object detection~\cite{andres2024ddeletion}. Yet deletion-based evaluation can itself introduce distribution shift through masking~\cite{hooker2019roar}, and existing studies have not examined whether detector explanation metrics are stable across driving domains or coupled to detector confidence.

    This work connects domain shift and evaluation confounds. Dataset bias is a long-standing concern~\cite{torralba2011bias}, and pedestrian detection is a clear case where domain changes affect model behavior~\cite{zhang2017citypersons}. Here, the question is whether explanation faithfulness, not only detection accuracy, is invariant under domain shift. Detector confidence is therefore treated as a potential confound to be controlled, rather than as a grouping variable to be interpreted naively.
	
	\section{Confidence-Controlled Auditing Protocol}
    \label{sec:method}

\begin{figure*}[!h]
      \centering
      \includegraphics[width=0.95\linewidth]{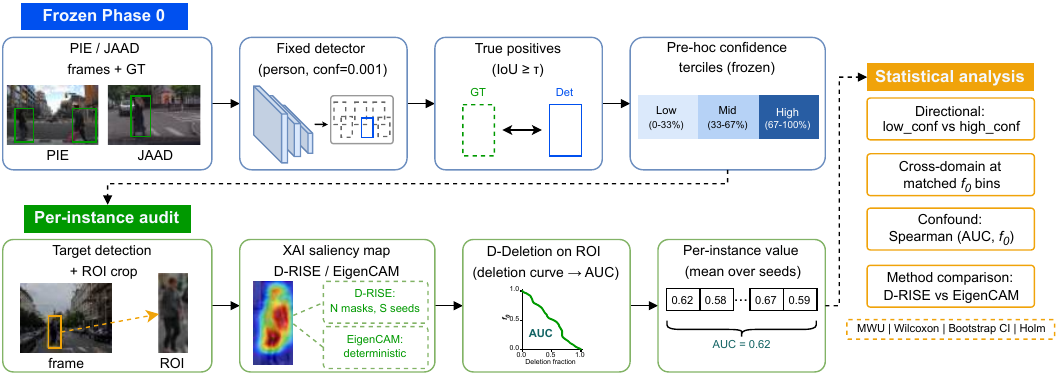}
      \caption{Overview of the confidence-controlled XAI auditing protocol.}
      \label{fig:pipeline}
    \end{figure*}

    The protocol audits whether the faithfulness of an explanation of a pedestrian detection is stable across driving domains, while treating the detector confidence as a variable to be controlled rather than one to be trusted. Its stages are summarized in Fig.~\ref{fig:pipeline}. The detector is kept fixed throughout, and each explanation method is applied under the same protocol, so that any difference reported below is attributable to the evaluation domain, or to the explanation method under a controlled comparison, and not to a change in the detector.

    \subsection{Setup and Notation}
    Let $\mathcal{D}$ be a fixed, pretrained detector that maps an image $x$ to a set of detections, each a triple $(b, c, s)$ of box, class, and confidence score. A ground-truth pedestrian $g$ with box $b_g$ in domain $\delta$ is treated as a detected instance when $\mathcal{D}(x)$ contains a detection of the person class with $\mathrm{IoU}(b, b_g) \geq \tau$; the confidence of that detection is denoted $s_g$. Ground-truth boxes below a minimum height are discarded as unreliable to audit. Only detected instances enter the study, since an explanation requires a detection to explain. For a target detection with box $b_0$ and class $c_0$, the detector response used throughout is the D-RISE target vector score~\cite{petsiuk2021drise}, which couples localization and category into a single scalar:
    
    \begin{equation}
    f(x) = \max_{\substack{(b,c,s)\in D(x)\\ c=c_0}} \operatorname{IoU}(b,b_0)\,s .
    \label{eq:targetvector}
    \end{equation}

    Because the maximization is restricted to detections predicted as $c_0$, $s$ is the confidence of the target-class detection; if no such detection is returned, $f(x)$ is set to zero.

    \subsection{Pre-hoc Confidence Stratification}
    Within each domain, instances are stratified into three confidence terciles. The tercile boundaries are the empirical $33.3$ and $66.7$ percentiles of the detection confidence over the instance pool of that domain, and they are computed and frozen before any explanation is produced. This ordering matters: fixing the stratification ahead of the explanations prevents the low-confidence tercile from being shaped by the very quantity later measured. The terciles are assigned from the full-frame detection confidence, which is the operational quantity of the deployed detector, whereas the explanations are computed on a region of interest, as described next.

    \subsection{Region-of-Interest Explanations}
    Computing the explanation over the full frame is both wasteful and misleading, since a pedestrian occupies a small fraction of a driving image and the deletion budget defined below would be spent almost entirely on background. For each target detection, a region of interest $R$ is therefore cropped around $b_0$, expanded by a fixed margin on each side, and all explanation and metric computations are carried out on $R$. Two explanation methods are audited under this common setup. The first is D-RISE~\cite{petsiuk2021drise}, a targeted, perturbation-based explainer. Its saliency map is obtained by sampling $N$ random low-resolution binary masks $M_i$, upsampling them to the size of $R$, and weighting each mask by the target response of the masked crop:
    
    \begin{equation}
    S \;=\; \frac{1}{N\,p}\sum_{i=1}^{N} f\!\left(R \odot M_i\right)\, M_i,
    \label{eq:drise}
    \end{equation}

    where $p$ is the probability that a mask cell is active and $\odot$ denotes elementwise masking. The second is EigenCAM~\cite{muhammad2020eigencam}, a class-agnostic baseline that projects the activations of a detector layer onto their first principal component. It is deterministic and requires a single forward pass, in contrast to the stochastic, many-pass D-RISE, which makes it a useful reference point for how much a targeted explainer gains under the same metric.

    \subsection{Detector Deletion Metric}
    Faithfulness is quantified with a deletion measure adapted to detection~\cite{andres2024ddeletion}. Pixels of $R$ are ranked by saliency $S$ in decreasing order and progressively removed toward a baseline value. At each deletion fraction the target response $f$ is recorded, yielding a deletion curve. The primary metric is the D-Deletion AUC, computed as the area under the deletion curve normalized by the deletion range, but not by the zero-deletion response. It is used for within-domain, paired-method, and $f_0$-matched cross-domain comparisons. A response-normalized variant is kept only as a diagnostic, since it becomes unstable when ROI re-detection is weak. Lower AUC indicates higher faithfulness.

    \subsection{Confidence Control}
    Because the deletion measure is computed from the target response, it can be coupled to the strength of the detection at explanation time. To separate a domain effect from this coupling, the detection strength at zero deletion, denoted $f_0$, is used as a control variable. Instances of both domains are grouped into fixed bins of $f_0$, and the two domains are compared within each bin. A difference that persists within matched $f_0$ bins is less likely to arise solely from one domain having weaker detections and is therefore treated as evidence of a domain-dependent faithfulness difference.

    \subsection{Statistical Protocol}
    Each perturbation-based explanation is stochastic through its masks, so every instance is evaluated over several seeds and summarized by the per-instance mean; the deterministic baseline is seed-invariant and contributes a single per-instance value after aggregation. Independent group comparisons use the Mann-Whitney U test, whereas paired method comparisons over common instances use the Wilcoxon signed-rank test. Effect sizes are reported as Cliff's delta, uncertainty as bootstrap 95\% confidence intervals of mean differences, and p-values within each comparison family are adjusted with the Holm correction. Medians, interquartile ranges, and sensitivity checks are used as descriptive robustness support; Section~\ref{sec:setup} lists all numerical parameters.

	\section{Experimental Setup}
    \label{sec:setup}

    \subsection{Datasets and Instance Sampling}
    The audit is carried out on two pedestrian datasets recorded from a moving vehicle, PIE~\cite{rasouli2019pie} and JAAD~\cite{rasouli2017jaad}, both at $1920\times1080$ resolution. To limit temporal redundancy, one frame is sampled every ten frames along each pedestrian track, and ground-truth boxes shorter than 30 pixels are discarded. From the resulting candidates, $6000$ instances per domain are sampled with a fixed seed for detector screening. Using IoU $\geq 0.5$ yields true-positive pools of $4386$ instances for PIE and $4747$ for JAAD. Confidence terciles are fitted on these pools and frozen before any explanation is produced. The audited evaluation set is then drawn from each pool, balanced across terciles, for a total of $300$ instances per domain, with $100$ per tercile.

    \subsection{Detector}
    The detector is YOLOv8s pretrained on COCO~\cite{jocher2023yolov8}, kept fixed for both domains. The confidence threshold at inference is set to $0.001$ rather than a conventional operating value. This choice is deliberate, since a higher threshold would remove precisely the weak detections that populate the low-confidence tercile. Only the person class is retained, and the input is processed at $640\times640$. Detections are taken after the detector's standard NMS; lowering the confidence threshold only enlarges the candidate set before true-positive filtering. Audited targets are always ground-truth-matched detections, and the response in (1) is computed as the maximum target similarity over post-NMS person detections.

    \subsection{Explanation and Metric Parameters}
    The perturbation-based explainer is D-RISE with $N=2000$ random masks generated on an $8\times8$ grid with cell activation probability $0.5$. The baseline is EigenCAM, computed from the activations of the last neck layer of the detector; it is deterministic and therefore seed-invariant. Every explanation is computed on a region of interest obtained by expanding the target box by its own size on every side, which yields a crop of about three times the box in each dimension. Before image-boundary clipping, this makes the target occupy approximately one ninth of the ROI area, reducing sensitivity to absolute pedestrian scale, although residual effects from clipping, resizing, and effective pedestrian size may remain. The deletion curve is evaluated at $25$ steps up to a maximum removed fraction of $0.5$ of the region, with removed pixels set to a zero baseline. This hard-deletion baseline is kept fixed across domains and methods; therefore, the analysis interprets relative differences rather than absolute AUC values. Faithfulness is summarized by the area under the deletion curve, as defined in Section~\ref{sec:method}.

    \subsection{Statistical and Confidence-Control Parameters}
    Each perturbation-based explanation is evaluated under five independent mask seeds and summarized by the per-instance mean. Confidence intervals are estimated by bootstrap with 10,000 resamples at the 95\% level, and p-values within each comparison family are adjusted with the Holm correction. The confidence-controlled analysis uses five $f_0$ bins with edges at 0.0, 0.2, 0.4, 0.6, 0.8, and 1.0; residual mean-$f_0$ differences between domains are inspected within each bin. Sensitivity checks repeat the controlled comparison with a response-normalized deletion variant and with alternative $f_0$ bins of width 0.25. Key parameters are listed in Table~\ref{tab:setup}. All experiments are run without training or fine-tuning on a single NVIDIA RTX 3060 with 6 GB of memory. Code and artifacts are available in the \href{https://github.com/rubendflorezzela/confidence-controlled-xai-auditing}{\textit{\textcolor{accent}{project repository}}}.

    \begin{table}[!h]
    \centering
    \caption{Experimental parameters.}
    \label{tab:setup}
    \footnotesize
    \renewcommand{\arraystretch}{1.08}
    \setlength{\tabcolsep}{3.5pt}
    \begin{tabularx}{\columnwidth}{@{}>{\raggedright\arraybackslash\bfseries}p{0.34\columnwidth}X@{}}
        \toprule
        \rowcolor{blue!10}[0pt][0pt]
        \textbf{Component} & \textbf{Configuration} \\
        \midrule

        \rowcolor{gray!12}[0pt][0pt]
        \multicolumn{2}{@{}l}{\textbf{Datasets and sampling}} \\
        Domains & PIE, JAAD; $1920\times1080$ frames \\
        Sampling & 1 frame every 10; min. GT height 30 px \\
        TP criterion / pool & IoU $\geq0.5$; 4386 PIE and 4747 JAAD \\
        Evaluation set & 300 per domain, balanced across terciles \\
        Terciles & 33.3 / 66.7 percentiles, frozen pre-hoc \\

        \midrule
        \rowcolor{gray!12}[0pt][0pt]
        \multicolumn{2}{@{}l}{\textbf{Detector and explainers}} \\
        Detector & YOLOv8s pretrained on COCO, fixed \\
        Inference & conf. 0.001; person class; input $640\times640$ \\
        D-RISE & 2000 masks, $8\times8$ grid, $p=0.5$, 5 seeds \\
        EigenCAM & last neck layer, deterministic \\

        \midrule
        \rowcolor{gray!12}[0pt][0pt]
        \multicolumn{2}{@{}l}{\textbf{Deletion, statistics, and hardware}} \\
        ROI / deletion & margin 1.0 ($\approx3\times$ box); 25 steps; max frac. 0.5 \\
        Statistics & 10{,}000 bootstrap samples; Holm correction \\
        $f_0$ bins & 0.0, 0.2, 0.4, 0.6, 0.8, 1.0 \\
        Hardware & NVIDIA RTX 3060, 6 GB \\
        Sensitivity checks & normalized AUC; $f_0$ bins of width 0.25 \\

        \bottomrule
    \end{tabularx}
\end{table}
	
	\section{Results}
    \label{sec:results}

    \subsection{Faithfulness Across Domains and Confidence Terciles}
    Table~\ref{tab:groups} reports D-Deletion by domain and tercile for D-RISE. In both domains the measure rises monotonically from the low- to the high-confidence tercile, from $0.083$ to $0.200$ in PIE and from $0.039$ to $0.161$ in JAAD. Since a higher value denotes a less faithful explanation, a direct reading would suggest that stronger detections receive less faithful explanations. The lower tercile boundary differs markedly between domains ($0.284$ for PIE versus $0.164$ for JAAD), whereas the upper boundary is similar ($0.687$ versus $0.682$), indicating that the largest confidence shift occurs in the lower part of the distribution. Both observations call for caution before any causal reading, as the next result makes explicit. The same ordering is observed in the medians, with interquartile ranges showing that the pattern is not driven only by a few high-AUC instances.

    \begin{table}[!h]
    \centering
    \caption{D-Deletion AUC by domain and confidence tercile for D-RISE (mean with $95\%$ bootstrap interval; higher is less faithful). $f_0$ is the mean detection strength.}
    \label{tab:groups}
    \begin{tabular}{llccc}
    \toprule
    \textbf{Domain} & \textbf{Tercile} & $n$ & \textbf{D-Deletion} & $f_0$ \\
    \midrule
    PIE  & low  & 100 & 0.083 [0.069, 0.097] & 0.338 \\
    PIE  & mid  & 100 & 0.125 [0.113, 0.138] & 0.577 \\
    PIE  & high & 100 & 0.200 [0.189, 0.210] & 0.742 \\
    \midrule
    JAAD & low  & 100 & 0.039 [0.030, 0.048] & 0.229 \\
    JAAD & mid  & 100 & 0.063 [0.052, 0.074] & 0.434 \\
    JAAD & high & 100 & 0.161 [0.147, 0.174] & 0.742 \\
    \bottomrule
    \end{tabular}
    \end{table}

    \subsection{Faithfulness Is Confounded With Detection Strength}
    \label{sec:results-B}
    The Spearman correlation between D-Deletion and the detection strength $f_0$ is high and positive, at $0.76$ over all instances, $0.82$ in JAAD and $0.70$ in PIE, each with $p<0.001$. 
    
\begin{figure}[!h]
        \centering
        \includegraphics[width=0.85\columnwidth]{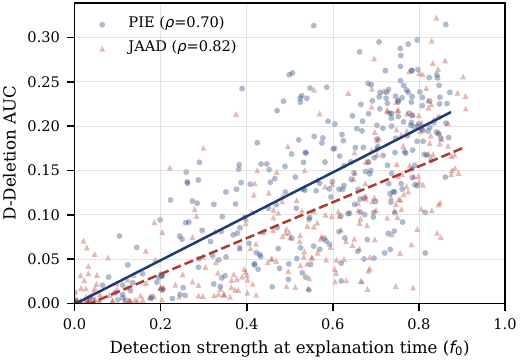}
        \caption{Detection strength at explanation time ($f_0$) versus D-Deletion for each audited instance, by domain. Higher \mbox{D-Deletion} denotes a less faithful explanation. The measure increases with $f_0$ in both domains, which is the confound, while PIE lies above JAAD at comparable $f_0$.}
        \label{fig:scatter}
    \end{figure}

    The coupling persists within individual terciles, where the confidence range is narrow, reaching $0.80$ and $0.73$ in the low-confidence terciles of JAAD and PIE. Fig.~\ref{fig:scatter} shows this dependence and, at the same time, a vertical separation between the two domains. Importantly, the strongest coupling is with $f_0$, rather than merely with the frame-level confidence used to define the terciles. Within the low-confidence tercile, frame-level confidence is essentially uncorrelated with \mbox{D-Deletion}, whereas $f_0$ remains strongly correlated. The raw tercile trend in Table~\ref{tab:groups} is therefore entangled with detection strength and cannot be attributed to explanation faithfulness alone.

    \subsection{Confidence-Controlled Cross-Domain Analysis}

    A direct low- versus high-confidence comparison is not interpreted as a safety-relevant finding. Under the raw AUC, low-confidence instances show smaller areas mainly because they have lower $f_0$, whereas the response-normalized diagnostic becomes unstable when ROI re-detection is weak. The directional axis is therefore reported as confounded, and the analysis focuses on comparisons matched by $f_0$.
    
\begin{figure}[!h]
        \centering
        \includegraphics[width=0.85\columnwidth]{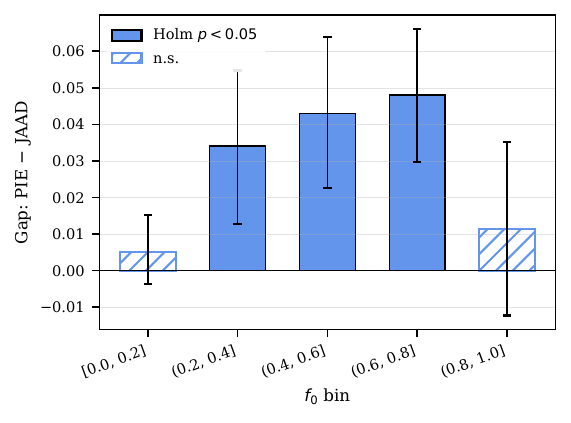}
        \caption{Cross-domain D-Deletion gap, PIE minus JAAD, within bins of matched detection strength $f_0$, with $95\%$ bootstrap intervals. Filled bars are significant after Holm correction; hatched bars are not.}
        \label{fig:bins}
    \end{figure}

    Grouping instances into $f_0$ bins controls the dominant score-related confound. For D-RISE, PIE shows higher \mbox{D-Deletion} than JAAD in the three central bins, with gaps of 0.034, 0.043, and 0.048 over $0.2 < f_0 \leq 0.8$, confidence intervals excluding zero, Cliff's delta between 0.35 and 0.42, and Holm-adjusted $p < 0.003$. The two extreme bins are not significant, consistent with smaller sample size and floor or ceiling effects. Mean $f_0$ values (PIE/JAAD) across the five matched bins were 0.089/0.059, 0.304/0.317, 0.510/0.503, 0.712/0.706, and 0.834/0.845, corresponding to absolute gaps of 0.030, 0.013, 0.007, 0.006, and 0.011. Residual imbalance was smallest in the three central bins where the D-RISE effect was resolved. Fig.~\ref{fig:bins} summarizes the D-RISE gaps.

    Table~\ref{tab:controlled} reports the same confidence-controlled analysis for both explainers. EigenCAM shows a weaker and less consistent pattern: the gap is positive in most bins, including the two upper-central bins, but no bin survives Holm correction. Two sensitivity checks support the D-RISE finding. The response-normalized deletion variant preserves the \mbox{PIE--JAAD} gap in the three central $f_0$ bins, although it becomes unstable in the lowest-response bin, and alternative $f_0$ bins of width 0.25 lead to the same qualitative conclusion outside the lowest-response region.

    \begin{table}[!h]
        \centering
        \caption{Cross-domain D-Deletion gap (PIE $-$ JAAD) within matched $f_0$ bins, per method. Positive gaps indicate PIE less faithful. $\delta$ is Cliff's delta; $p_H$ is the Holm-adjusted $p$-value. Significant gaps ($p_H<0.05$) in bold.}
        \label{tab:controlled}
        \footnotesize
        \setlength{\tabcolsep}{3.2pt}
        \begin{tabular}{llccccc}
        \toprule
        \textbf{Method} & \textbf{$f_0$ bin} & \textbf{Gap} & \textbf{95\% CI} & $\delta$ & $p_H$ & $n_{\text{P}}/n_{\text{J}}$ \\
        \midrule
        D-RISE & [0.0, 0.2] & 0.005 & [$-$0.004, 0.015] & 0.16 & 0.382 & 30/76 \\
        D-RISE & (0.2, 0.4] & \textbf{0.034} & [0.013, 0.055] & 0.42 & \textbf{0.002} & 47/40 \\
        D-RISE & (0.4, 0.6] & \textbf{0.043} & [0.023, 0.063] & 0.35 & \textbf{0.002} & 69/61 \\
        D-RISE & (0.6, 0.8] & \textbf{0.048} & [0.030, 0.066] & 0.40 & \textbf{$<$0.001} & 122/81 \\
        D-RISE & (0.8, 1.0] & 0.011 & [$-$0.013, 0.034] & 0.14 & 0.382 & 32/42 \\
        \midrule
        EigenCAM & [0.0, 0.2] & 0.022 & [0.000, 0.044] & 0.24 & 0.231 & 30/76 \\
        EigenCAM & (0.2, 0.4] & $-$0.003 & [$-$0.047, 0.040] & 0.01 & 1.000 & 47/40 \\
        EigenCAM & (0.4, 0.6] & 0.040 & [$-$0.012, 0.092] & 0.17 & 0.299 & 69/61 \\
        EigenCAM & (0.6, 0.8] & 0.061 & [0.002, 0.120] & 0.19 & 0.115 & 122/81 \\
        EigenCAM & (0.8, 1.0] & $-$0.008 & [$-$0.119, 0.104] & 0.01 & 1.000 & 32/42 \\
        \bottomrule
        \end{tabular}
    \end{table}

    \subsection{Comparison With a Non-Perturbative Baseline}
    The same protocol is applied to EigenCAM, a class-agnostic map that requires a single forward pass and no random masking, and its saliency map is seed-invariant. 
    Across the 600 instances common to both explainers, D-RISE attains a markedly lower D-Deletion AUC than EigenCAM, with mean AUC 0.112 versus 0.350 and a median paired difference of -0.179 (Wilcoxon signed-rank test, \mbox{$p<0.001$}).
    
    Under the ROI-based deletion metric, therefore, the targeted perturbation-based explanations are substantially more faithful than the non-perturbative baseline. The confidence coupling of Section~\ref{sec:results-B} also appears for EigenCAM, with an overall Spearman correlation of 0.74 and a value up to 0.92 within a low-confidence tercile. This suggests that the coupling is not specific to D-RISE and is related to the deletion-based evaluation setup. The cross-domain pattern is weaker and less consistent for EigenCAM: the gap is positive in most bins, including the two upper-central bins, but no bin remains significant after Holm correction. Thus, only D-RISE provides statistically resolved evidence of the controlled domain effect, whereas EigenCAM serves as a less faithful non-perturbative baseline under the same deletion protocol.
    
\begin{figure*}[!h]
    \centering
    \includegraphics[width=0.85\linewidth]{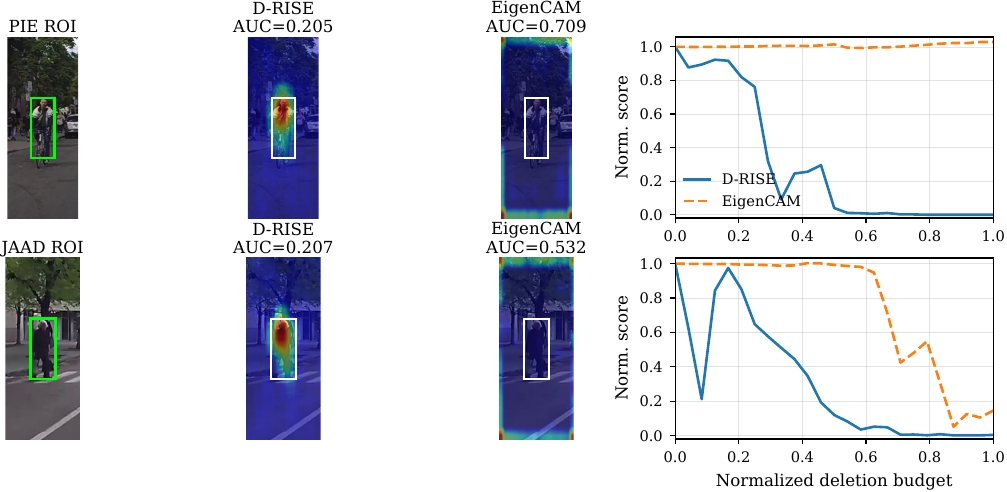}
    \caption{Representative PIE and JAAD examples from the same $f_0$ bin. D-RISE saliency is averaged over five mask seeds. Curves are normalized by the zero-deletion response for visualization; AUC labels report raw ROI-based D-Deletion for the displayed maps (lower is more faithful).}
    \label{fig:qualitative}
    \end{figure*}

    Fig.~\ref{fig:qualitative} qualitatively compares representative examples from the $0.6 < f_0 \leq 0.8$ bin: D-RISE concentrates saliency on the pedestrian and yields lower raw ROI-based D-Deletion AUC, whereas EigenCAM spreads saliency over broader context and retains higher deletion responses.

   \section{Discussion and Conclusion}

    The results have a methodological and an empirical reading. Methodologically, D-Deletion is not independent of the detection strength at explanation time. The Spearman correlations of 0.70 to 0.82 for D-RISE, together with the comparable EigenCAM coupling, show that confidence-stratified faithfulness comparisons are confounded unless $f_0$ is controlled. This does not invalidate deletion metrics, but it supports their use with detection-strength control in safety-oriented audits.

    Empirically, within matched $f_0$ bins, D-RISE faithfulness is not invariant across domains: PIE yields higher D-Deletion than JAAD over the central $f_0$ range, with moderate effect sizes and Holm-adjusted significance. EigenCAM shows a weaker and less consistent version of this pattern and does not remain significant after correction. The gap is therefore reported as an observed domain-dependent faithfulness difference, not as evidence that one dataset is intrinsically better.

    Practically, the reported D-Deletion gaps should not be interpreted as direct safety margins. They indicate that the same detector--explainer pair can yield different faithfulness estimates across driving domains even at matched detection strength. Explanation quality should therefore be validated per operating domain before saliency maps are used for debugging, monitoring, or safety-case evidence.

    Several limitations remain. The deletion metric inherits the sensitivity of masking-based evaluation to out-of-distribution artifacts, mitigated here by relative rather than absolute interpretation. The study uses one fixed detector, two datasets, and two explainers, so the magnitude of the effects may vary with other architectures or domains. Terciles are assigned from full-frame confidence, whereas explanations are computed on ROI crops; object scale, ROI occupancy, occlusion, and illumination may therefore act as secondary confounds. Descriptive stratification by pedestrian size and occlusion was also inspected, but small strata limited these checks to a diagnostic role. The extreme $f_0$ bins are also underpowered, and $f_0$ remains an imperfect proxy for visual difficulty. Future work should extend the audit to further detectors, datasets, and detection-specific explanation methods.

    In summary, this work introduced a confidence-controlled protocol for auditing pedestrian-detection explanations across driving domains. Applied to PIE and JAAD with a fixed YOLOv8s detector, D-RISE, and EigenCAM, it shows that deletion-based faithfulness is coupled to detection strength and that, after controlling this strength, D-RISE faithfulness is domain-dependent over the central $f_0$ range. D-RISE is best interpreted as an offline or periodic audit, whereas EigenCAM provides a single-pass baseline; real-time onboard use would require lighter explainers or reduced masking budgets.
	
	
	\bibliographystyle{IEEEtran}
	\bibliography{root} 

@inproceedings{redmon2016yolo,
  author={Redmon, Joseph and Divvala, Santosh and Girshick, Ross and Farhadi, Ali},
  booktitle={2016 IEEE Conference on Computer Vision and Pattern Recognition (CVPR)}, 
  title={You Only Look Once: Unified, Real-Time Object Detection}, 
  year={2016},
  volume={},
  number={},
  pages={779-788},
  doi={10.1109/CVPR.2016.91}
}

@inproceedings{rasouli2019pie,
  author={Rasouli, Amir and Kotseruba, Iuliia and Kunic, Toni and Tsotsos, John},
  booktitle={2019 IEEE/CVF International Conference on Computer Vision (ICCV)}, 
  title={{PIE}: A Large-Scale Dataset and Models for Pedestrian Intention Estimation and Trajectory Prediction}, 
  year={2019},
  volume={},
  number={},
  pages={6261-6270},
  doi={10.1109/ICCV.2019.00636}
}

@inproceedings{rasouli2017jaad,
  author={Rasouli, Amir and Kotseruba, Iuliia and Tsotsos, John K.},
  booktitle={2017 IEEE International Conference on Computer Vision Workshops (ICCVW)}, 
  title={Are They Going to Cross? A Benchmark Dataset and Baseline for Pedestrian Crosswalk Behavior}, 
  year={2017},
  volume={},
  number={},
  pages={206-213},
  doi={10.1109/ICCVW.2017.33}
}

@article{atakishiyev2024xai,
  author={Atakishiyev, Shahin and Salameh, Mohammad and Yao, Hengshuai and Goebel, Randy},
  journal={IEEE Access}, 
  title={Explainable Artificial Intelligence for Autonomous Driving: A Comprehensive Overview and Field Guide for Future Research Directions}, 
  year={2024},
  volume={12},
  number={},
  pages={101603-101625},
  doi={10.1109/ACCESS.2024.3431437}
}

@article{omeiza2022explanations,
  author={Omeiza, Daniel and Webb, Helena and Jirotka, Marina and Kunze, Lars},
  journal={IEEE Transactions on Intelligent Transportation Systems}, 
  title={Explanations in Autonomous Driving: A Survey}, 
  year={2022},
  volume={23},
  number={8},
  pages={10142-10162},
  doi={10.1109/TITS.2021.3122865}
}

@imisc{petsiuk2018rise,
  title={{RISE}: Randomized Input Sampling for Explanation of Black-box Models}, 
  author={Vitali Petsiuk and Abir Das and Kate Saenko},
  year={2018},
  eprint={1806.07421},
  archivePrefix={arXiv},
  primaryClass={cs.CV},
  url={https://arxiv.org/abs/1806.07421}, 
}

@inproceedings{petsiuk2021drise,
    author    = {Petsiuk, Vitali and Jain, Rajiv and Manjunatha, Varun and Morariu, Vlad I. and Mehra, Ashutosh and Ordonez, Vicente and Saenko, Kate},
    title     = {Black-Box Explanation of Object Detectors via Saliency Maps},
    booktitle = {Proceedings of the IEEE/CVF Conference on Computer Vision and Pattern Recognition (CVPR)},
    month     = {June},
    year      = {2021},
    pages     = {11443-11452}
}

@article{andres2024ddeletion,
    title = {On the black-box explainability of object detection models for safe and trustworthy industrial applications},
    journal = {Results in Engineering},
    volume = {24},
    pages = {103498},
    year = {2024},
    issn = {2590-1230},
    doi = {https://doi.org/10.1016/j.rineng.2024.103498},
    url = {https://www.sciencedirect.com/science/article/pii/S259012302401750X},
    author = {Alain Andres and Aitor Martinez-Seras and Ibai Laña and Javier {Del Ser}},
}

@article{dollar2012ped,
  author={Dollar, Piotr and Wojek, Christian and Schiele, Bernt and Perona, Pietro},
  journal={IEEE Transactions on Pattern Analysis and Machine Intelligence}, 
  title={{Pedestrian Detection}: An Evaluation of the State of the Art}, 
  year={2012},
  volume={34},
  number={4},
  pages={743-761},
  doi={10.1109/TPAMI.2011.155}
}

@inproceedings{zhang2017citypersons,
    author = {Zhang, Shanshan and Benenson, Rodrigo and Schiele, Bernt},
    title = {{CityPersons}: A Diverse Dataset for Pedestrian Detection},
    booktitle = {Proceedings of the IEEE Conference on Computer Vision and Pattern Recognition (CVPR)},
    month = {July},
    year = {2017}
}

@inproceedings{selvaraju2017gradcam,
    author = {Selvaraju, Ramprasaath R. and Cogswell, Michael and Das, Abhishek and Vedantam, Ramakrishna and Parikh, Devi and Batra, Dhruv},
    title = {{Grad-CAM}: Visual Explanations From Deep Networks via Gradient-Based Localization},
    booktitle = {Proceedings of the IEEE International Conference on Computer Vision (ICCV)},
    pages={618--626},
    month = {Oct},
    year = {2017}
}

@inproceedings{hooker2019roar,
    author = {Hooker, Sara and Erhan, Dumitru and Kindermans, Pieter-Jan and Kim, Been},
    title = {A benchmark for interpretability methods in deep neural networks},
    year = {2019},
    publisher = {Curran Associates Inc.},
    booktitle = {Proceedings of the 33rd International Conference on Neural Information Processing Systems},
    articleno = {873},
    numpages = {12}
}

@inproceedings{torralba2011bias,
  author={Torralba, Antonio and Efros, Alexei A.},
  booktitle={CVPR 2011}, 
  title={Unbiased look at dataset bias}, 
  year={2011},
  volume={},
  number={},
  pages={1521-1528},
  doi={10.1109/CVPR.2011.5995347}
}

@software{jocher2023yolov8,
  author={Jocher, Glenn and Chaurasia, Ayush and Qiu, Jing},
  title={Ultralytics {YOLOv8}},
  version={8.0.0}, 
  year={2023},
  url={https://github.com/ultralytics/ultralytics}
}

@inproceedings{muhammad2020eigencam,
  author={Muhammad, Mohammed Bany and Yeasin, Mohammed},
  booktitle={2020 International Joint Conference on Neural Networks (IJCNN)}, 
  title={{Eigen-CAM}: Class Activation Map using Principal Components}, 
  year={2020},
  volume={},
  number={},
  pages={1-7},
  doi={10.1109/IJCNN48605.2020.9206626}
}
	
\end{document}